\pdfoutput=1
\documentclass[11pt]{article}

\usepackage[]{EMNLP2023}
\usepackage{graphicx} 
\usepackage{graphics} 
\usepackage{epsfig}
\usepackage{booktabs}
\usepackage{times}
\usepackage{latexsym}
\usepackage{amssymb}
\usepackage{amsmath,amsthm,amssymb,amsfonts}
\usepackage{makecell}
\usepackage{multirow}
\usepackage{xcolor}
\usepackage{colortbl}  
\usepackage{hyperref}
\usepackage{booktabs}       % professional-quality tables

\usepackage[T1]{fontenc}
\usepackage[utf8]{inputenc}

\usepackage{microtype}

\usepackage{inconsolata}

\title{Prototype-based HyperAdapter for Sample-Efficient Multi-task Tuning}

\newcommand*\samethanks[1][\value{footnote}]{\footnotemark[#1]}

\author{
\small
    Hao Zhao\textsuperscript{1}\begin{NoHyper}\thanks{~~Equal technical contribution, co-first authors.}\end{NoHyper}\quad 
    Jie Fu\textsuperscript{2}\samethanks[1]~~\begin{NoHyper}\thanks{~~Corresponding authors.}\end{NoHyper}\quad 
    Zhaofeng He\textsuperscript{1}\samethanks[2]
\\
\small
    \textsuperscript{1}Beijing University of Posts and Telecommunications\quad \textsuperscript{2}Hong Kong University of Science and Technology \quad 
\\
\small
\texttt{\{haozhao,zhaofenghe\}@bupt.edu.cn} \quad 
\texttt{jiefu@ust.hk} 
}

\begin{document}
\maketitle

\begin{abstract}
Parameter-efficient fine-tuning has shown its effectiveness in adapting the pre-trained language models to downstream tasks while only updating a small number of parameters. 
Despite the success, most existing methods independently adapt to each task without considering knowledge transfer between tasks and are limited to low-data regimes. 
To overcome this issue, we propose Prototype-based HyperAdapter (PHA), a novel framework built on the adapter-tuning and hypernetwork. 
It introduces an instance-dense retriever and a prototypical hypernetwork to generate the conditional modules in a sample-efficient manner.
This leads to comparable performance improvements against existing Parameter-efficient fine-tuning methods on multi-task learning and few-shot transfer learning. 
More importantly, when the available data size gets smaller, our method outperforms other strong baselines by a large margin. 
Based on our extensive empirical experiments across various datasets, we demonstrate that PHA strikes a better trade-off between trainable parameters, accuracy on stream tasks, and sample efficiency. Our code is publicly available at \url{https://github.com/Bumble666/PHA} 

\end{abstract}

\section{Introduction}

Fine-tuning a pre-trained language model (PLM) yields extraordinary potential for simultaneous adaptation to multiple downstream tasks in a multi-task setting. However, fine-tuning all the parameters of models induces substantial storage and deployment costs, especially as pre-trained model sizes are growing rapidly. To address this issue, several works \citep{adapter,lester-etal-2021-power, NEURIPS2021_081be9fd,hu2022lora,ding2022delta,gui-xiao-2023-hifi,zeng-etal-2023-one,liao-etal-2023-parameter,xie-lukasiewicz-2023-empirical} have developed parameter-efficient fine-tuning which trains compact modules per task and adapts PLMs to downstream tasks. Nonetheless, these methods require learning different modules to adapt to diverse tasks, and the cost of the parameter increases proportionally with the number of tasks. On the other hand, training task-specific modules separately fails to reap benefits from other relative tasks. 
\begin{figure}
\centering
\includegraphics[width=1.0\columnwidth]{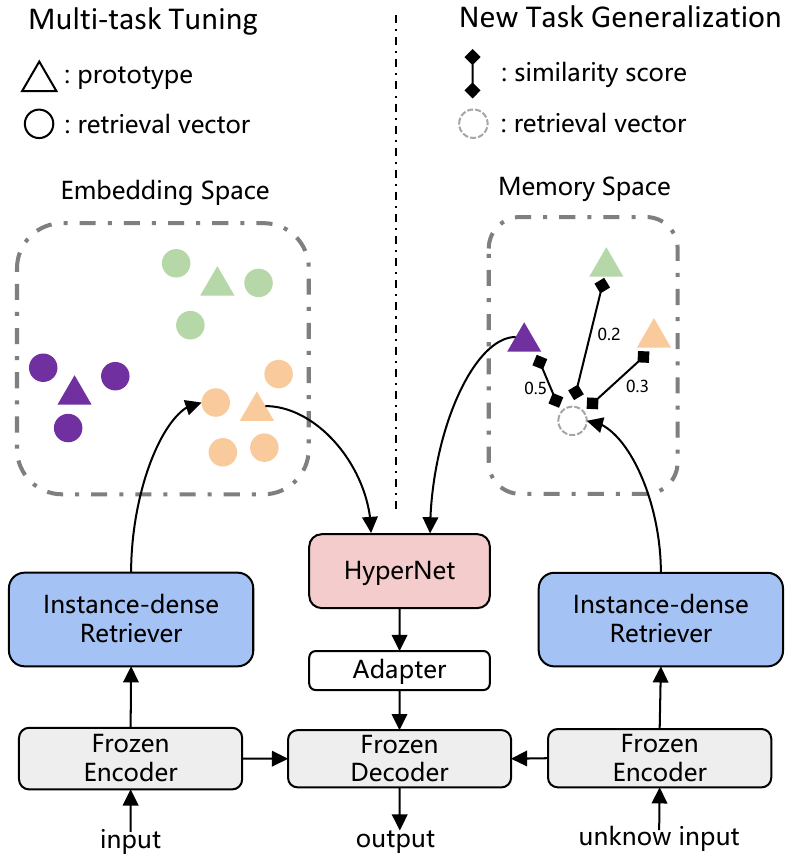}
\caption{Prototype-based HyperAdapter generates task-specific adapter with a shared hypernetwork and prototypes calculated by task-specific instances. During generalization, the relevant task prototype is retrieved by the Instance-dense Retriever, and the new adapter layers are generated by hypernetwork.}
\label{fig1}
\end{figure}

Recent work \citep{hyperformer,hyperdecoder} has proposed training a hypernetwork to generate the parameters of these modules to achieve a better trade-off between parameter efficiency and adaption for downstream tasks. These methods encourage the multi-task learning model to capture the shared information by leveraging task-shared hypernetwork while eliminating negative task interference by generating conditioned modules individually. Despite these methods' success in multi-task learning, there are still some issues:(1) hypernetwork-based methods generally optimize the specific embedding and shared hypernetwork together by end-to-end training without any regularization. Task-specific information is inseparably intertwined, which suppresses the efficiency of the hypernetwork, especially in resource-limited settings. (2) these existing approaches generalize to new tasks that require task-specific prior knowledge or knowledge from frozen pre-trained models.

These works \citep{hyperformer,pfeiffer2023modular} indicate that the task-shared hypernetwork serves as a cross-task information captor, while a specific embedding should encapsulate task-level semantic features in order to extract pertinent information from the hypernetwork for generating corresponding module parameters. Empirically, task-level features are typically implicitly represented by related instance features. A natural idea to encourage embedding generation is to calculate the central points (prototypes) of task-specific instance features.

In this paper, we introduce the Prototype-based HyperAdapter(PHA), a novel framework built on adapter-tuning that achieves both multi-task learning and generalization to new tasks in a sample-efficient manner. As depicted in Figure~\ref{fig1}, PHA consists of two main components, Instance-dense Retriever and Prototypical HyperNetworks. The first part aims to train a retriever to discriminate the instances from different tasks in embedding space. For the second part, we aim to estimate the task-specific prototypes with the instance-level features and keep the prototypes as embeddings to be trained with the hypernetwork.

Specifically, we project the encoded instance features into embedding space using the retriever. To avoid instances interference in embedding space, we train the retriever with the InfoNCE estimator \citep{oord2018representation}. As a result, it clusters intra-task instances and increases the distances between inter-task instances. The projected features here can be deemed as instance-level semantic features that are used to estimate task-level embeddings.  
Inspired by PCL \citep{li2021prototypical}, we estimate the task-specific embedding using the contrastive prototypical loss, which encourages the prototypes to become the center points of instance-level features. Compared with the existing method, where the specific embeddings are optimized directly during tuning, our method efficiently learns specific embedding with side information, which helps to optimize the embedding space in low-data regimes. During the adaptation of new tasks, since we maintain the previous task-level semantic features as prototypes that align with the instances in the embedding space, we match the corresponding prototype for the current new task by calculating the distance between the new instances and the previous prototypes.

We evaluate PHA on 13 NLP datasets across diverse tasks.  Extensive experiments show the effectiveness of PHA, especially in low-data regimes. Meanwhile, PHA is able to achieve a few-shot domain adaption with 4-32 shots. For example, PHA outperforms the strong multitask adapter transfer baseline by 1.0\% with lower trainable parameters on the GLUE benchmark. In low resource regimes where only 100 samples per task from the GLUE benchmark is available, PHA outperforms adapter-tuning by 8.0\%. Our analysis shows that PHA efficiently captures specific information and shared information while reducing negative transfer. We also present a detailed analysis to demonstrate that the instances from different tasks can be identified by corresponding prototypes and used for new task adaption. 

\section{Background}
% We start by introducing the required background on the adapter tuning and hypernetwork.
\noindent\textbf{HyperNetworks.} A hypernetwork\citep{hypernetworks} can generate parameters to be used by networks or modules. Specifically, the hypernetwork, denoted as $h_w$, leverages an embedding $\boldsymbol{I}$ to generate the module parameters $\boldsymbol{\phi}$:	
\begin{equation}
\boldsymbol{\phi} = h_w(\boldsymbol{I})
\end{equation}

\noindent\textbf{Adapter.} Adapter\citep{adapter} is commonly used in parameter-efficient tuning that aims to apply PLM to downstream tasks. Specifically, adapter-based tuning inserts trainable task-specific modules into transformer layers, while keeping the PLM fixed. The adapter $A^l(x)$ for layer $l$ is defined as:
\begin{equation}
A^l(\boldsymbol{x})=\boldsymbol{D^l}(\text{ReLU} ((\boldsymbol{U^l}({x})))+x,
\end{equation}
where $\boldsymbol{D^l} \in \mathbb{R}^{d \times b }$ and $\boldsymbol{U^l} \in \mathbb{R}^{b \times d }$ are down/up-projection matrices. $x \in \mathbb{R}^d$ refers to the input. $d$ is the hidden dimension of PLM, $b$ is the bottleneck size satisfying $b \ll d$.

\citet{he2021towards} propose using the parallel adapter, which is different from the traditional sequential insertion method. They demonstrate the more efficient parameter-efficient fine-tuning that the parallel adapter offers.

\section{Method}
\textbf{Problem Setup.} We are following a general multi-task learning problem. Given a pre-trained language model ${\mathcal{M}_{\theta}}$ with parameters $\theta$ and a set of target tasks $\{D\}=\left\{\mathcal{D}_{1}, \mathcal{D}_{2}, \ldots, \mathcal{D}_{\tau}\right\}$, where $\tau$ is the total number of tasks and $\left\{\mathcal{D}_{i}\right\}=\left\{x_{i}^{n}, y_{i}^{n}\right\}_{n=1}^{N_{i}}$ represents the training data of the $i$-th task with $N_{i}$ samples. 
The main objective of our study is to fine-tune ${\mathcal{M}_{\theta}}$ for downstream tasks $\{D\}$ using a multi-task learning setup and to ensure that it is capable of generalizing to new tasks.

\noindent\textbf{Method Overview.} The key idea of our approach is to directly learn prototype embeddings for each task from training instances which is acquired by using a task-shared encoder, then generating task-specific adapter layers by retrieving prototype embedding and feeding it into a hypernetworks. As shown in Figure~\ref{fig1}, the encoded instances are projected into retrieval vectors by the instance-dense retriever for prototype learning. These prototypes that represent the task-specific information enable hypernetwork to update efficiently. This potentially allows more sample-efficient fine-tuning and few-shot transfer learning. 

\subsection{Instance-dense Retriever}
We define an instance-dense retriever for better generalization. Let $h_i \in \mathbb{R}^{d}$ denote the last layer's mean-pooled hidden state of the training sample, which belongs to the $i$-th task. To align the embedding feature with the training sample in the latent space, an instance-dense retriever $G(\cdot)$ is applied to construct the retrieval vector $z_i=G(h_i)$, where $G(\cdot)$ is an MLP consisting of two feed-forward layers and a ReLU non-linearity. Furthermore, we need the retriever to have the ability which explicitly encourage alignment between instances from the same task, as well as push away instances from different tasks.

To efficiently learn the discriminative retriever, we introduce the following loss function $\mathcal{L}_{IR}$ based on the InfoNCE:
\begin{equation}
\mathcal{L}^{i}\!=\!\sum_{z_i \in D}\frac{-1}{N_i-1} \!\sum_{z_j \in \hat{D}_i} \!\log \!\frac{\exp f({z}_{i}\cdot {z}_{j})}{\sum\limits_{z_{m}\in S(i)} \exp f({z}_{i}\cdot {z}_{m})},
\end{equation}

\begin{equation}
\mathcal{L}_{\mathrm{IR}}\!=\frac{1}{\tau} \sum_{i=1}^{\tau} \mathcal{L}^{i},
\end{equation}
where $\mathcal{L}^{i}$ is the learning objective for task $i$, $f(\cdot)$ is the cosine similarity function. ${\hat{D}_i}$ is a set of positive samples of $z_i$ and $S(i)$ denotes a set of negative samples for $z_i$.

The instance-dense retriever aggregates instance-level information from the same task and enables flexible reuse of knowledge used for few-shot transfer learning.

\subsection{Prototypical HyperNetworks}
Simply using the learned task embedding to encapsulate task-specific information biases the task-shared hypernetwork to overfit the training data distribution, which means that inadequate sample efficiency and mixed knowledge are more susceptible to changes in distribution during cross-task transferring. 

To overcome this issue, we propose to implicitly exploit the instance-level information to instruct the task embedding instead of end-to-end training. To be more specific, we found that contrastive formulation is an efficient strategy for learning robust and sample-efficient Hypernetworks. We first initialize a set of embedding $\{k_i\}_{i=1}^{\tau}$, where $\{k_i\} \in \mathbb{R}^{d}$ is a trainable vector to learn the specific information of the $i$-th task. 
The learning objective for an embedding is
\begin{equation}
\mathcal{L}^{i}=\sum_{z_i \in D}\frac{-1}{N_i-1} \log \frac{\exp f({z}_{i}\cdot {k}_{i})}{\sum\limits_{k_{m}\in V(i)} \exp f({z}_{i}\cdot {k}_{m})},
\label{eq5}
\end{equation}
\begin{equation}
\mathcal{L}_{\mathrm{Pro}}=\frac{1}{\tau} \sum_{i=1}^{\tau} \mathcal{L}^{i},
\end{equation}
where $V(i)$ is a set of negative embedding. The objective forces each embedding to make use of the relative relationships between samples across tasks and avoid sample-inefficient knowledge transfer.

To generate specific parameters for different transformer layers and reduce the number of trainable parameters, we introduce a learnable layer embedding denoted as $e_m$, following a similar recipe as in Hyperformer \citep{hyperformer}. $m$ denotes the $m$-th layer of transformer model.

Let $H(\cdot)$ denote the HyperNetwork which generates the weight matrices $D_i^m$ and $U_i^m$ for task conditional adapter $A_i^m$: 
\begin{equation}
(D_i^m, U_i^m)=H(C(k_i, e_m)),
\end{equation}
where $C(\cdot)$ is a project network to concatenate the task embedding and layer embedding into a mixed embedding $I_{i}^{m}$.

Inspired by \citet{pfeiffer-etal-2021-adapterfusion} and \citet{he2021towards}, we only insert these conditional parameters (Adapter) into the Feed-Forward Networks (FFN) sub-layer in parallel: 
\begin{equation}
\mathbf{y}=\mathrm{FFN}({LN}(\mathbf{x}))+\text {A}(LN(\mathbf{x})),
\end{equation}
where $LN(\cdot)$ represents the LayerNorm layer. This enables efficient decoupling of knowledge from different tasks to task prototypes and adapts the changeable data distribution during transfer learning.
\subsection{Multi-task Tuning and New Task Generalization}
PHA achieves sample-efficient multi-task learning and few-shot adaption with different training methods.

\noindent\textbf{Multi-task Tuning.} We follow a general multi-task learning setup, where the task identity is included, and the different datasets are concatenated together. To achieve efficient fine-tuning, the encoder with task-shared adapters is used for encoding the training sample, and we estimate the embedding corresponding to the retrieval vector given the task identity via the loss function in Equation~\ref{eq5}. The decoder is attached by the specific adapters conditioned on contextual information. The whole model is trained in a sequence-to-sequence setting with the following objective function:
\begin{equation}
\mathcal{L}_{\mathrm{Total}}=\mathcal{L}_{\mathrm{PLM}}+\lambda(\mathcal{L}_{\mathrm{IR}}+\mathcal{L}_{\mathrm{Pro}}),
\label{gongshi9}
\end{equation}
where $\mathcal{L}_{\mathrm{PLM}}=\sum_{i=1}^{\tau}\mathcal{L}_i$ denotes the cross-entropy loss for all training tasks and $\lambda$ is a scalar balancing factor.

PHA allows the specific embedding to efficiently capture contextual information which helps the hypernetwork to generate the parameters of adapter layers in sample-efficient adaption.

\noindent\textbf{New Task Generalization.} For few-shot adaption, we retrieve the adaption embedding $k_a$ by calculating the similarity scores of retrieval vector $z$ and learned embeddings $\{k\}$ after multi-task training:
\begin{equation}
 k_a=\underset{i}{\arg \max } \, f(k_{i} \mid z).
 \label{gongshi10}
\end{equation}
% The retrieved adaption embedding is the most relevant to the target embedding due to the 
During training, we feed the adaption embedding to hypernetworks that generate the weight matrices of adapters for new tasks and optimize with the cross-entropy loss. 

Our method enables efficient generalization to new tasks with limited training examples, owing to the retrieved adaption embedding containing knowledge similar to that required for the new task.

\section{Experiments}

\subsection{Datasets}
Following prior works on multi-task learning for natural language understanding (NLU) tasks, we consider 8 datasets from GLUE \citep{wang-etal-2018-glue} benchmark and 4 datasets from SuperGLUE~\citep{NEURIPS2019_4496bf24} benchmark to evaluate the performance of our models. This benchmarks is a collection of text classification tasks including CoLA~\citep{Cola} for sentence acceptability, SST-2~\citep{sst2} for sentiment analysis, MNLI~\citep{mnli}, QNLI~\citep{qnli}, RTE~\citep{rte}, CB~\citep{de2019commitmentbank} for natural language inference, STS-B~\citep{sstb} for sentence similarity, MRPC~\citep{mrpc}, QQP~\citep{glue} for paraphrasing similarity, WSC~\citep{wsc} for coreference resolution, BoolQ~\citep{clark-etal-2019-boolq} for question answering and WiC~\citep{wic} for word sense disambiguation. In addition, we also introduced an additional dataset: SciTail~\citep{Khot_Sabharwal_Clark_2018} for few-shot adaption.

\begin{table*}
\centering
\resizebox{\textwidth}{!}{
\begin{tabular}{l|c|cccccccc>{\columncolor{gray!40}}c|cccc>{\columncolor{gray!40}}c}
\toprule 
\quad & \quad &  \multicolumn{9}{c|}{\textbf{GLUE}} & \multicolumn{5}{c}{\textbf{SuperGLUE}}\\
\midrule
\textbf{Method} & \textbf{\makecell{Tunable\\Params}} & \textbf{CoLA}& \textbf{SST-2}& \textbf{STS-B}& \textbf{MRPC}& \textbf{QQP}& \textbf{MNLI}& \textbf{QNLI}& \textbf{RTE}& \textbf{Avg} &\textbf{BoolQ}&\textbf{WiC} &\textbf{CB} &\textbf{WSC} &\textbf{Avg}\\
\midrule
{$\text{FT} ^{\dagger}$} & {220M} & {\textbf{61.8}}& {\textbf{94.6}}& {89.7}& {90.2}& {\textbf{91.6}}& {\textbf{86.8}}& {93.0}& {71.9}& {84.9}&\textbf{{81.1}}	&\textbf{{70.2}}	&{85.7}&	{59.6}	&{74.2}
\\

{$\text{Shared-FT} ^{\dagger}$} & {28M} & {54.9}& {92.5}& {88.8}& {90.2}& {91.1}& {85.7}& {92.0}& {75.4}& {83.8}&{78.5}&	{69.5}	&{85.2}	&{66.7}	&{75.0}
\\

{$\text{Adapter} ^{\dagger}$} & {1.9M} & {64.0}& {93.2}& {\textbf{90.7}}& {85.3}& {90.2}& {86.5}& {93.2}& {71.9}& {84.5}&{82.5}	&{67.1}	&{85.7}	&\textbf{67.3}	&{75.7}
\\

{$\text{Shared-Adapter} ^{\dagger}$} & {1.8M} & {61.5}& {93.0}& {89.9}& {90.2}& {90.5}& {86.3}& {93.2}& {70.3}& {84.4}&{78.4}	&{67.3}&{85.2}	&{64.7}	&{73.9}
\\
{$\text{PT} ^{\ast}$} & {76.8k} & {10.6}& {90.9}& {89.5}& {68.1}& {89.7}& {81.3}& {92.8}& {54.7}& {72.2}&61.7	&48.9	&67.9	&51.9	&57.6
\\

{$\text{Hyperformer++} ^{\dagger}$} & {638K} & {59.0}& {93.7}& {90.3}& {88.6}& {89.9}& {85.0}& {93.3}& {\textbf{77.5}}& {84.7}&{75.8}	&{68.9}	&{81.5}	&{52.9}&	{69.8}
\\
{$\text{HyperDecoder} ^{\ddagger}$} & {1.8M} & {55.9}& {94.0}& {90.5}& {87.7}& {90.5}& {86.0}& {93.4}& {71.7}& {83.7}&77.8&	66&	92.6&	66.7&	75.8
\\
{PHA (ours)} & {616K} & {60.6}& {94.0}& {88.9}& {{89.2}}& {90.9}& {86.3}& {\textbf{93.4}}& \textbf{80.4}& {\textbf{85.5}}&80.7&	64.8&	\textbf{96.3}&	62.7&	\textbf{76.1}
\\
\bottomrule
\end{tabular}
}
\caption{\label{t1}
Overall comparison on Multi-task Adaptation. {T5-base} is used as the PLM backbone of all methods. We also report \textbf{Tunable Params}, which represents the number of parameters that need to be fine-tuned for each task. The best result on each block is in \textbf{bold}. For GLUE results, $\dagger$  denotes results reported from~\citep{hyperformer}. $\ddagger$ denotes results reported from~\citep{hyperdecoder}. $\ast $ denotes results reported from~\citep{attempt}.
}

\end{table*}
\subsection{Baselines}
To evaluate the effectiveness of our proposed method, we conduct an analysis against several established methods that serve as strong baselines for multi-task learning: \textbf{Adapter} \citep{adapter} and \textbf{Shared-Adapter}, we train adapters on a single task or a group of tasks and place them into transformer layers. \textbf{Hyperformer} \citep{hyperformer} and \textbf{Hyperdecoder} \citep{hyperdecoder} that use task-conditioned or sample-conditioned Hypernetworks to generate adapters and place them into transformer layers. In addition, We compare our method with \textbf{Fully fine-tuning(FT)}, and \textbf{Shared-FT} share the model across different tasks. We also compare the state-of-art prompt transfer methods: \textbf{Prompt tuning(PT)} \citep{lester-etal-2021-power}, prompt tuning prepends tunable embeddings to the input layer, and the embeddings are initialized with each task respectively. \textbf{SPoT} \citep{spot} and \textbf{ATTEMPT} \citep{attempt}, \textbf{MPT} \citep{wang2023multitask} adapts to the target tasks with the shared prompts obtained by distilling knowledge from the source tasks.

\subsection{Experiments Details}
\label{sec53}
Following the setting of \citet{hyperformer}, when an original testing set is unavailable, the validation set is utilized as the testing set. In situations where the dataset contains less than 100k records, the validation set is divided into two sets: validation and testing. Conversely, larger datasets utilize 1000 training set samples selected for validation, with the original validation set used for testing. For the multi-task adaptation experiment, we performed multi-task learning on 8 datasets from GLUE and 4 datasets from SuperGLUE. For the low-data adaption experiment, we separately sample each individual task in GLUE with different proportions and quantities $(100,500,1000,2000,4000,1\%,3\%,5\%)$. As for the evaluation strategy, we use Pearson Correlation for STS-B and accuracy for other tasks as metrics. We save a checkpoint every 1000 steps for all models and report the average performance of all tasks on a single checkpoint.
In the few-shot adaption experiment, we randomly sample $k=4,16,32$ instances from the training set while the entire test set is used for testing. 
We mainly use T5-Base (220M) model \citep{raffel2019exploring} as the pre-trained language model. In addition, we also use T5-Small (60M) and T5-Large (770M) to explore the effect of model size on PHA performance in Section~\ref{sec544}. Unless specified, we train for 65k steps using the AdamW \citep{loshchilov2018decoupled} optimizer and set the batch size as 128 for all experiments. During training, the initial learning rate is set to 3e-4 with linear decay and 500 warm-up steps. We set the balancing factor $\lambda=0.1$ in Eq.~\ref{gongshi9} and keep it fixed for all our experiments. 
All experiments run for 5 times with different seeds and we report the average for each result. The detailed configurations per method on diverse datasets are shown in Appendix~\ref{A1}.

\begin{table*}
\setlength\tabcolsep{0.6cm}
\centering
\resizebox{\textwidth}{!}{
% \begingroup
% \setlength{\tabcolsep}{6pt} % Default value: 6pt
% \renewcommand{\arraystretch}{1} % Default value: 1

\begin{tabular}{l|c|cccccccc|c}
\toprule
\multicolumn{2}{c|}{k-shot} & {$\text{FT} ^{\dagger}$} & {$\text{Adapter} ^{\dagger}$} & {$\text{PT} ^{\dagger}$} & {$\text{SPoT} ^{\dagger}$} & {$\text{HF} ^{\dagger}$} & {$\text{ATP} ^{\dagger}$} & HD & {$\text{MPT} ^{\ddagger}$} & PHA(our) \\
\midrule
\multirow{3}{*}{4}   & CB      & 57.7	&51.1	&53.5&	71.4	&60.7	&\textbf{82.1}&	69.1&73.6& 76.5
         \\
& SciTail     & 79.6&	79.5&	57.7&	69.6&	82.0&	80.2	&75.4	&80.2&	\textbf{82.5}
         \\
& BoolQ    & 50.5&	53.4	&61.6	&50.5	&48.0	&61.8&	54.4&62.2&	\textbf{68.2}
        \\
\midrule
\multirow{3}{*}{16}      & CB      & 77.0	&74.8	&63.5	&64.3	&76.3	&78.5	&75.3	&78.6&\textbf{79.6}
         \\
& SciTail     & 80.0	&83.2	&60.8	&71.9	&86.5	&79.5	&85.4	&87.3&\textbf{87.7}
         \\
& BoolQ     & 56.5&	51.4	&61.9&	50.6&	50.2&	60.0&	64.6	&63.3&\textbf{71.3}
         \\ 
\midrule
\multirow{3}{*}{32} & CB      & 80.0&	74.8	&67.8	&64.3	&81.4&	\textbf{85.7}&	79.6&82.1&	82.7
         \\
& SciTail     & 81.9	&85.0	&60.2	&71.9&	85.8	&80.2&	85.1	&86.3&\textbf{88.6}
         \\
& BoolQ     &58.4	&54.5	&61.7&	61.2	&58.3&	65.3&	68.3&68.9&	\textbf{72.0}
   \\
\bottomrule
\end{tabular}
% \endgroup
}

\caption{\label{t2}
Result for few-shot transfer ($k={4,16,32}$). We report the accuracy for all tasks over 10 random seeds. The best result on each block is in \textbf{bold}. All of the models are trained on GLUE tasks with T5-Base backbone. $\dagger$ denotes results reported from~\citep{attempt}. $\ddagger$ denotes results reported from~\citep{wang2023multitask}. 
}
\end{table*}

\subsection{Results and Analysis}

\subsubsection{Multi-task Adaptation}
Table~\ref{t1} shows the evaluation results on GLUE and SuperGLUE. The results indicate that PHA outperforms all comparative methods regarding performance improvement while maintaining parameter efficiency.
Note that we do not compare with SPoT, ATTEMPT, and MPT since they require pre-training prompts to save the knowledge from source tasks and transfer them to target tasks. Extending these methods to the same setting where only pre-trained models are available is beyond our scope. Therefore, under the experimental setup of multi-task learning, our method cannot achieve a fair comparison with them. It is worth mentioning that MPT, ATTEMPT, and our method both use the same two-step training method in the few-shot transfer setting (Section~\ref{sec433}).
% It lies beyond our scope to extend these methods to the same setting where only pre-trained models are available. 
Specifically, our PHA approach achieves a performance increase of $+1.0\%$ over Adapter while only using $3 \times$ fewer trainable parameters. When we compared with the state-of-art adapter-based multi-task methods including recent Hyperformer++ and Hyperdecoder that use a hypernetwork to generate conditioned parameters similar to our approach, PHA achieves 0.8\% and 1.8\% point accuracy improvements respectively on GLUE while utilizing the same or lower number of trainable parameters. This demonstrates the potential of our method to reduce the negative interference between tasks better.
In addition, we observe that FT performs the best in all experimental methods, while it requires 220M trainable parameters on a single task. We find that our PHA is as competitive as that of the FT(85.5 vs. 84.9) and reduces the trainable parameters from $100\%$ to $0.28\%$. We also analyze the effectiveness of PHA in parameter efficiency, as detailed in Appendix~\ref{B1}.

\begin{figure}
\centering
\includegraphics[width=1.0\columnwidth]{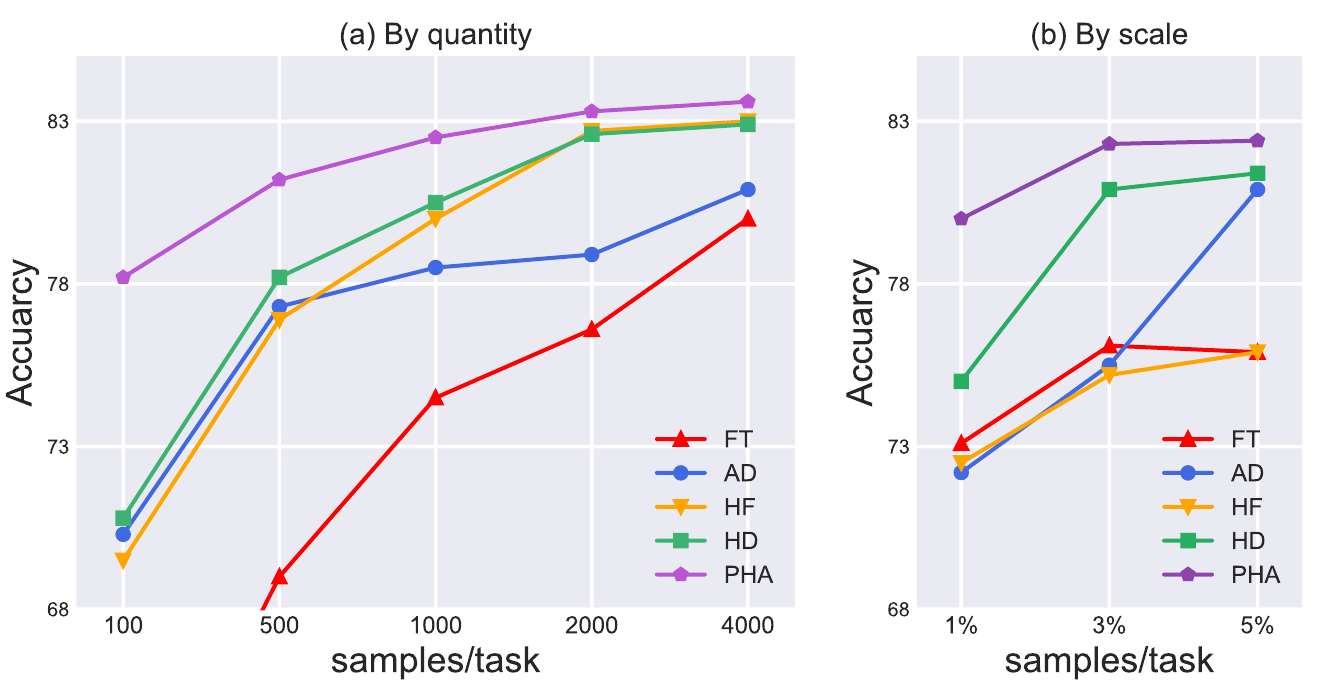}
\caption{Results on GLUE for sample efficiency experiments, which sample the various number or proportions of training samples per task.
AD, HF and HD denote Adapter, HyperFormer and HyperDecoder.}
\label{extra2}
\end{figure}

\subsubsection{Low-data Adaptation}
In our framework, we posit that the task prototypes, which are estimated by instance-level features, can better represent task information and improve the performance of hypernetworks in a sample-efficient manner during multi-task adaptation. To verify that our proposed method generalizes better when there are only limited available resources, we conduct low-data adaption experiments on the GLUE benchmark. Following \citet{hyperformer}, we train all models(Fine-tuning, Adapter, Hyperformer++, and Hyperdecoder) for 15k steps. More details are in Section~\ref{sec53}.

As shown in Figure~\ref{extra2}, PHA outperforms other baselines consistently across different configurations. We obverse that Fine-tuning performs the worst in cases of limited available data. A potential reason is that tuning all parameters of the base model makes it susceptible to keep the model in the state of over-fitting, especially in the low-data regimes. As for other experimental methods, we observe that when the number of available samples is relatively smaller, the existing state-of-the-art multi-task methods with hypernetworks close to the performance of Adapter until the number of available samples increases. This observation indicates that hypernetworks struggle to generate optimal parameters when using randomly initialized task embeddings or instances as contextual information under low-data regimes. Moreover, to better simulate the training task distribution in the GLUE benchmark, we randomly sample each individual task in GLUE for different proportions. Figure~\ref{extra2}(b) shows the comparison between PHA and adapter-based methods. Similar to the results in Figure~\ref{extra2}(a), our proposed method outperforms others by a large margin. We also conduct experiments on 4 datasets (BoolQ, WiC, CB, WSC) from SuperGLUE, as detailed in Appendix~\ref{c1}.

The superior performance of PHA over all competing methods indicates that our proposed task prototype efficiently captures contextual information to enable hypernetwork to generate the parameters of adapter layers.

\subsubsection{Few-shot Adaptation}
\label{sec433}

We explore how our proposed method performs when adapting to new tasks with sample-efficient. Specifically, following \citet{wang2023multitask}, we conduct few-shot experiments on BoolQ, CB, SciTail and compare PHA with the strong baselines, including Fine-tuning, Adapter, Prompt tuning, SPoT, Hyperformer, ATTEMPT, HyperDecoder, and MPT. The results in Table~\ref{t2} are obtained by training an 8-task adaptation for GLUE and fine-tuned with few-shot samples from BoolQ, CB, and SciTail. More details are in Section~\ref{sec53}.

Table~\ref{t2} summarizes the results on the few-shot adaption setting. Among Adapter-based transfer methods, PHA brings around $3 \% \sim 20 \%$ absolute improvements over the Adapter across different settings. While Hyperformer achieves better generation for new tasks, it requires us to have a precise understanding of target tasks. PHA significantly improves the performance of Hyperformer without requiring task-specific prior knowledge. In addition, our proposed method significantly outperforms other prompt-based transfer methods in most settings.

Our results demonstrate that our approach effectively generalizes to new tasks despite the limited availability of training samples.

\begin{table}
\centering
% \resizebox{\textwidth}{!}{
\begin{tabular}{cc|c}
\toprule
{\textbf{Prototype}} & \textbf{Retriever} & \textbf{GLUE Score}\\
\midrule
$\times $ & $\times $  &84.0\\
$\times $ & $\surd $&	84.3\\

$\surd $ & $\times$&	84.7\\

$\surd $ & $\surd $&	\textbf{85.5}\\
\bottomrule
\end{tabular}
% }
\caption{\label{t6}
Ablation study on GLUE.
}

\end{table}

\begin{figure}
\centering
\includegraphics[width=1.0\columnwidth]{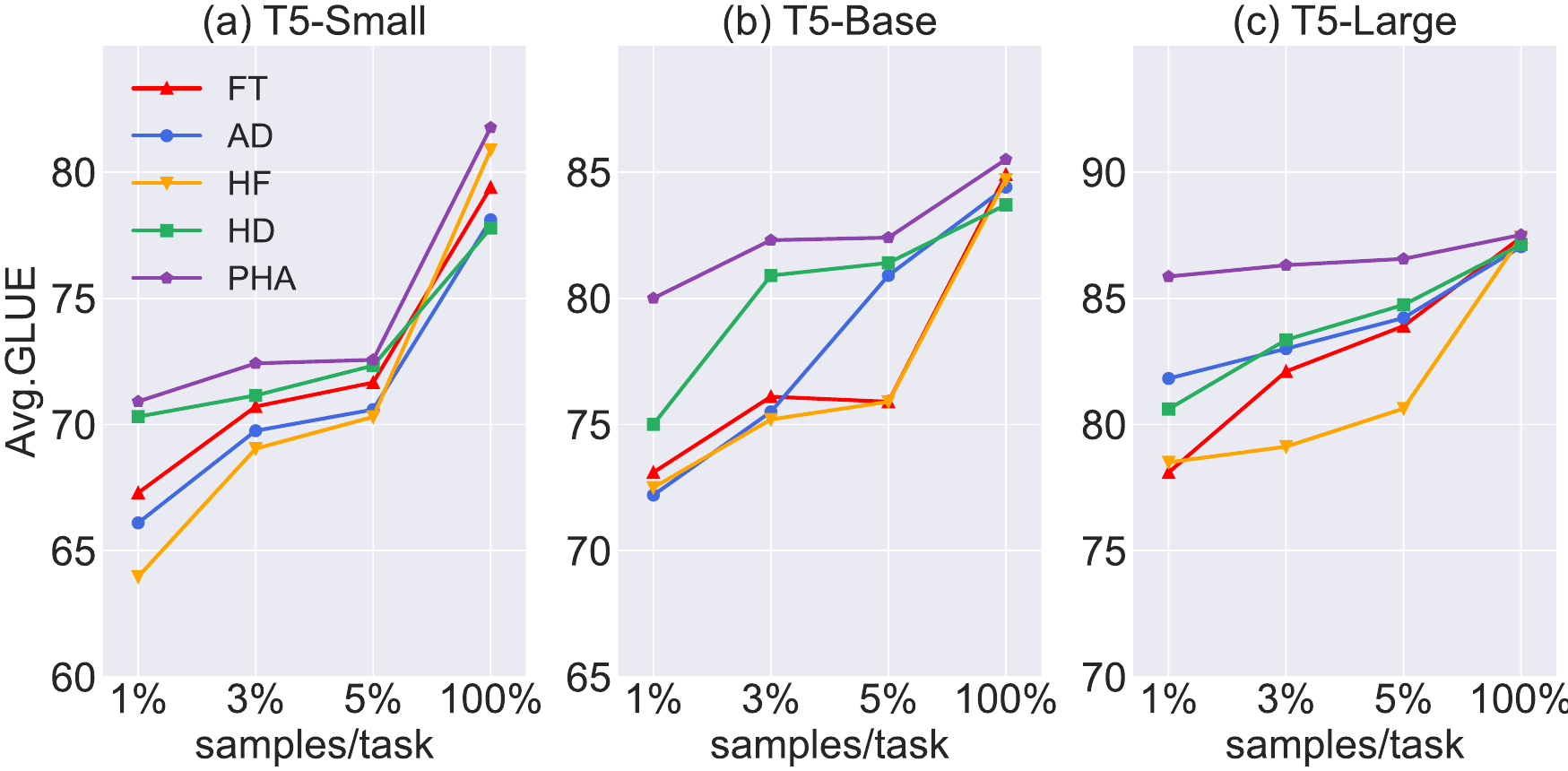}

\caption{The comparison of PEFTs (Adapter, Hyperformer++, Hyperdecoder, and our proposed PHA) and FT on GLUE with three scale pre-trained models under low-data regimes and full-data regimes.}
\label{fig5}
\end{figure}
\subsubsection{Ablation Study}
\begin{figure}
\centering
\includegraphics[width=1.0\columnwidth]{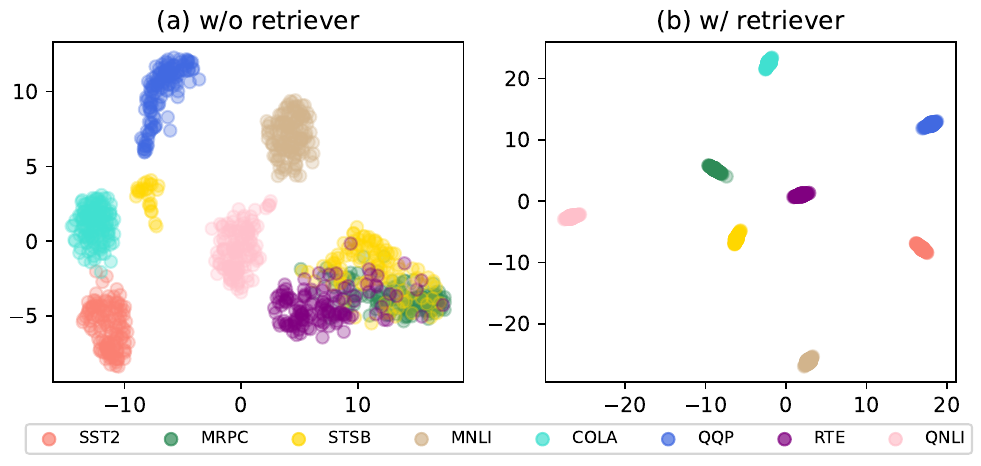}
\caption{t-SNE visualizations for samples w/ (right) and w/o (left) retriever.}
\label{fig2}
\end{figure}

We perform an ablation study on the GLUE benchmark to evaluate the effectiveness of the proposed modules. The prototype design and the instance-dense retriever are removed independently for this purpose.
As shown in Table~\ref{t6} (row 2), when we remove the prototype design and use the retrieval instances to train, we observe the performance has a significant drop. This shows that the instance-level information hinders the positive transfer across tasks under the limitation of hypernetwork capacity, while our design of task prototypes allows hypernetwork to capture shared-task information well, which is vital for enabling positive transfer across tasks. Table~\ref{t6} (row 3) removes the retriever. The task prototypes are estimated by the originally encoded instances. This results in intertwined task prototypes due to the relative dispersion of instance information in the embedding space. The decrease in performance suggests that adding the instance-dense retriever enables the prototype to encode task-specific knowledge better. Furthermore, we provide a visualization of the encoded instances from the GLUE benchmark to compare the effect of adding and removing the instance-dense retriever, as shown in Figure~\ref{fig2}. While samples belonging to the same task tend to be located near each other in the latent space, samples from different classes (e.g., STS-B, MRPC, RTE) still interleave with each other. After the retriever is added, instances from the same task are tightly clustered, while different tasks are widely separated.

\subsubsection{Impact of Model Scale}
\label{sec544}
To verify that our method is applicable to different pre-trained model sizes, we also experiment T5 with sizes from Small (60M) to Large (770M) on GLUE datasets, while reporting the average result of PHA as well as fully Fine-tuning (FT), Adapter (AD), Hyperformer++ (HF) and Hyperdecoder(HD). As shown in Figure~\ref{fig5}, under three scales of pre-trained model, we find that PHA achieves superior and competitive performances in low-data and full-data regimes, respectively. 
% In addition, PHA 
This indicates that our proposed prototypical strategy is still able to achieve the best sample efficiency when the size of large-scale transformer models increases.

\subsubsection{Effect of Retriever for Generalization.}
\begin{figure}
\centering
\includegraphics[width=1.0\columnwidth]{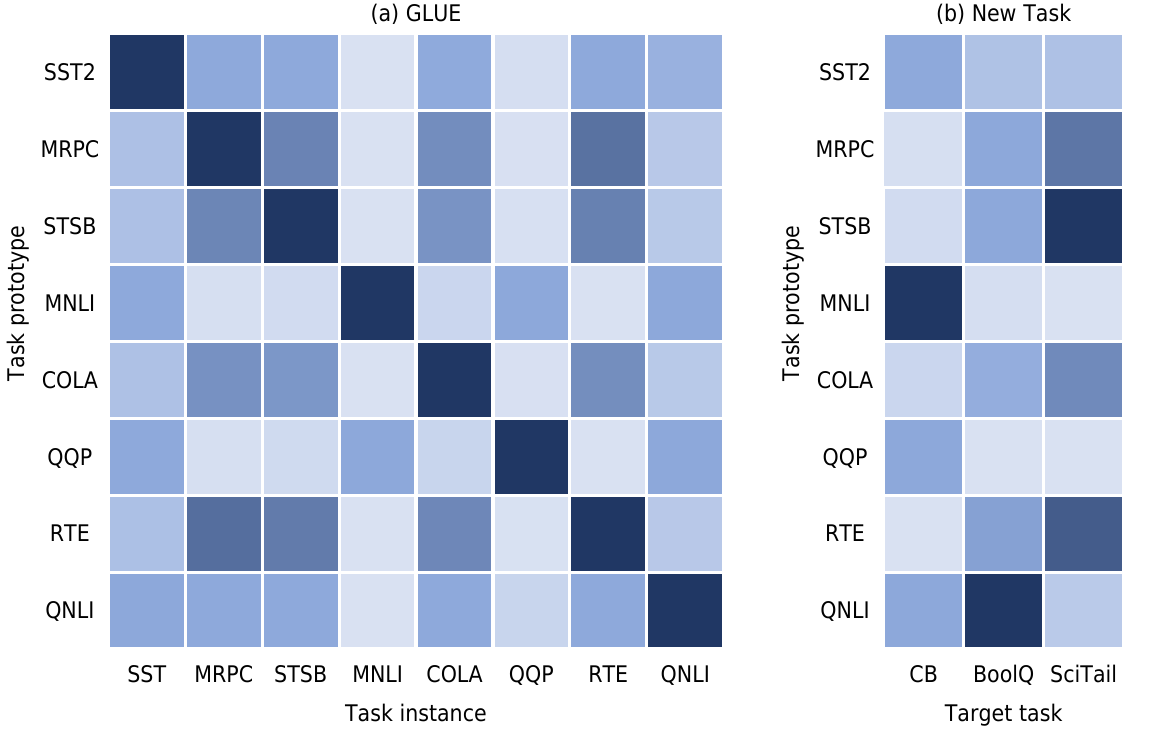}
\caption{Visualization of the similarity scores, which are calculated by the dense retriever, between the instances and task prototype in GLUE (a) and between pre-trained prototypes and the task-agnostic instances in three datasets (b). Darker colors indicate higher scores.}
\label{fig3}
\end{figure}

The retriever plays an important role in adapting to new tasks. To explore how the retriever works after training under a multi-task setting, we consider the similarity scores, described in Eq.~\ref{gongshi10}, to measure retrieval results. Specifically, we randomly sample each individual task in GLUE and calculate the similarity scores through the trained task prototypes and retrieval vectors transferred by the trained retriever. Figure~\ref{fig3}(a) shows a visualization of similarity scores. We find that the retriever precisely retrieved the task identity of the corresponding task instance. This suggests that task prototypes and instance vector has aligned in embedding space to enable more efficient capture of single-task common features. 

We also demonstrate that the retriever has the ability to match the corresponding task prototype to target tasks that require generalization. Figure~\ref{fig3}(b) illustrates that the similarity score is relatively high for related tasks such as CB and MNLI, SciTail, and STSB, all of which belong to the NLI task family. As for QNLI and BoolQ, since the task prototype trained on GLUE does not include Boolean Question Answering (QA) tasks, the retriever has matched QNLI prototype, which is in the same domain as BoolQ. Therefore, our proposed method can naturally generalize to new tasks when the related task prototype and the hypernetwork containing cross-task knowledge are both available.

\section{Related Work}
\textbf{Multi-task Learning and Transfer.}
Multi-task learning(MTL) aims to take advantage of the shared information between the different tasks and train a unified model to simultaneously solve multiple tasks. 
In the context of NLP, this is typically achieved by sharing certain layers across all tasks while using task-specific layers for specific tasks~\citep{liu-etal-2019-multi}. With the popularization of large language models (LLMs), \citet {raffel2019exploring} explores the training of LLMs on various tasks which are transformed into a unified format, and some works \citep{aghajanyan-etal-2021-muppet,aribandi2021ext5,sanh2022multitask,wei2022finetuned} indicate that the LLMs can be better generalized to new tasks through large-scale multitasking training. More recent work \citep{pfeiffer-etal-2021-adapterfusion,spot,attempt,wang2023multitask} focuses on multi-task transfer with parameter-efficient fine-tuning as the increasing LM size. Though the effectiveness of multitask learning is improved, they need to finetune the LLMs twice on source tasks, which are carefully selected, and multiple target tasks. This limits the applicability of the methods. Differently, our proposed method only requires a pre-trained model to achieve multi-task learning and transfer to a new task.

Several works~\citep{jin-etal-2020-language,hyperformer,hyperdecoder,pmlr-v162-he22f} introduce hypernetworks~\citep{hypernetworks} to share the cross-task information by generating the parameters of adapter layers~\citep{adapter} from specific embeddings during multi-task learning. Our work is motivated by~\citet{hyperdecoder}, but proposes to use task-level information represented by prototypes to optimize the embedding distribution of hypernetworks, which reduces negative transfer between different tasks and improves the performance of adaption across tasks, especially in low-resource regimes.

\noindent\textbf{Prototype Learning.}
Prototype learning is widely used to improve the representation ability of networks for few-shot learning. Some works~\citep{gao2019hybrid,NEURIPS2020_70feb62b,ding2021prototypical,li2021prototypical} indicate that the prototype is forced to learn some common features of samples within the class by prototype estimation. \citet{cui-etal-2022-prototypical} propose to construct a verbalizer for prompt-based few-shot tuning by estimating prototypes with contrastive learning. This differs from our method, which uses a prototypical strategy to explore the specific information for corresponding tasks.

\section{Conclusion}
We introduce Prototype-based HyperAdapter, a novel framework built on adapter-tuning. Our method achieves both multi-task adaption and adaption to a new task in a sample-efficient manner. It generates parameters of adapter layers conditioned on task-specific prototypes which are calculated by the corresponding instance-level features. In addition, the specific prototype is retrieved and transferred to new tasks to be further tuned.
The resulting method significantly outperforms previous SOTA on full/low-data multi-task adaption and few-shot adaption.

\section*{Limitations}
Our work has demonstrated strong experimental results and sample efficiency in multitasking adaption. However, there are several limitations:
Firstly, in few-shot adaption, the method we proposed, which tunes the base model on 8 NLP tasks, can generalize to new target tasks efficiently. But tuning on more large-scale tasks may result in better generalization improvements. Secondly, as shown in Figure~\ref{fig3}, a new task may be related to multiple task prototypes, rather than a single one. In our method, we only select the most relevant prototypes, which may ignore the transfer of some weakly related knowledge.
In addition, we use adapters in this work, but our method could possibly also benefit from other parameter-efficient approaches \citep{lester-etal-2021-power, mahabadi2021compacter,li-liang-2021-prefix, hu2022lora,liu-etal-2022-p}. 

\section*{Acknowledgements}
This work is supported by National Key R\&D Program of China
(Grant No. 2022YFF1202400), Major Science and Technology Innovation Program of Hangzhou (Grant No. 2022AIZD0154), Natural Science Foundation of China (Grant No. 62176025, 62301066), Beijing Nova Program (Grant No. 20220484161), the Fundamental Research Funds for the Central Universities (Grant No. 2023RC72) and Theme-based Research Scheme (T45-205/21-N), Research Grants Council of Hong Kong.

\bibliography{anthology,custom}
\bibliographystyle{acl_natbib}
\clearpage
\appendix

\section{Implementation Details}
\label{A1}
We set the dimension of the retrieval vector ($z_i$), the task prototype ($k_i$), and the layer embedding $z_m$ to 128. For the instance-dense retriever $G(\cdot)$ with bottleneck architecture, we select the bottleneck size to 128. For the mixed embedding $I_i^m$, the dimension of $I_i^m$ is selected as 32 for multi-task experiments and 64 for other experiments. We use the default hyperparameters by \citet{hyperdecoder} for Adapter and Hyperdecoder. The hyperparameters of other baselines are set according to the original papers~\citep{hyperformer,attempt,wang2023multitask}. Following \citet{hyperformer,hyperdecoder}, the length of the sequence is 128 at the encoder and the decoder. We use its HuggingFace~\citep{wolf-etal-2020-transformers} Pytorch~\citep{NEURIPS2019_bdbca288} implementation. All data is obtained from huggingface datasets and preprocessed into a ”seq2seq” format following \citet{raffel2019exploring}.
% \subsection{Low-data Experiments Details}
% \label{A12}
\section{Parameter Efficiency}
\label{B1}
Our proposed method not only balances sample efficiency and accuracy on stream tasks but also achieves parameter efficiency. 
We compare the trainable parameters of PHA with other baselines.
We assume that the basic model is a transformer model with an $L$-layer encoder-decoder structure for $\tau$-tasks and $d$ is the model dimension. The tasks-share adapter is attached to the encoder layer and the condition-generate adapter is attached to the decoder layer in our work. The bottleneck dimension of Adapter is $b$. Adapter for the single layer costs $2db+b+d$. Given bottleneck size $d^{\prime}$, the instance-dense retriever $G(\cdot)$ with bottleneck architecture to construct the retrieval vector $z \in \mathbb{R}^{d^{\prime}}$, which result in $d{d^\prime}+{d^\prime}^2$ parameters. We have the task prototype $k_i \in \mathbb{R}^{d^\prime}$ for $i$-th task and layer embedding $e_m \in \mathbb{R}^{d^\prime}$ for $m$-th layer. These have a total of $\tau {d^\prime}+ L{d^\prime}$ parameters. The project network $C(\cdot)$ is used for concatenating the task prototype $k_i \in \mathbb{R}^{d^\prime}$ and layer embedding $e_m \in \mathbb{R}^{d^\prime}$ into a mixed embedding $I_i^m \in \mathbb{R}^{d_h}$, which result in $2{d^\prime}d_h$ parameters. The hypernetworks costs ${d_h}(2db+b+d)$ parameters. The total cost of PHA is $d{d^\prime}+{d^\prime}^2+(\tau {d^\prime}+ L{d^\prime})+2{d^\prime}d_h+{d_h}(2db+b+d)+L(2db+b+d)$. 

There are a large number of tasks and transformer layers in our work settings. 
For the adapter, its parameter count increases linearly with the task and number of layers. For Hyperformer, its parameter count is more than our method due to using two hypernetworks. We used a bottleneck structured retriever to reduce trainable parameters compared to Hyperdecoder.

\section{Additional Results }
\label{c1}
Here, we conduct more experiments to demonstrate the sample efficiency of our method on different
datasets (BoolQ, WiC, CB, and WSC) with T5-Base pre-trained model. The result are included in Table~\ref{t9}. The performance PHA with
T5-Base achieves the best performance in terms of accuracy and sample efficiency in the 4 datasets from SuperGLUE. In addition, to verify the effectiveness of our method on NLU tasks, we conducted sample efficiency experiments on three datasets: Xsum, WMT16 Ro-En, and WMT16 En-Ro. Due to limited computation resources, we sampled 1\% of the data and conducted experiments on three backbone models of different scales(small, base, and large). The results are shown in Table~\ref{t10}.
\begin{table}
\centering
% \resizebox{\linewidth}{!}{
\scalebox{0.8}{
\begin{tabular}{c|ccc|c}
\toprule
{\textbf{\makecell{samples/\\task}}} & \textbf{AD} & \textbf{HF} & \textbf{HD} & \textbf{PHA}\\
\midrule
1\% & 57.8 &50.4&	57.7&	\textbf{59.2}\\
\midrule
3\% &61.3	&51.3&	59.6	&\textbf{67.6}\\
\midrule
5\% &60.3	&54.3	&59.9&	\textbf{69.3}\\
\bottomrule
\end{tabular}
% }
}
\caption{\label{t9}
Results on 4 datasets (BoolQ, WiC, CB, WSC) from SuperGLUE for the various proportions of training samples per task (1\%,3\%,5\%). We report the average accuracy. AD, HF and HD denote Adapter, HyperFormer and HyperDecoder.
}
\end{table}

\begin{table}
\centering
% \resizebox{\linewidth}{!}{
\scalebox{0.8}{
% \begingroup
% \setlength{\tabcolsep}{6pt} % Default value: 6pt
% \renewcommand{\arraystretch}{1} % Default value: 1

\begin{tabular}{l|c|ccc|c}
\toprule
\multicolumn{2}{c|}{Method} & {$\text{XSUM} $} & {$\text{Ro-En} $} & {$\text{En-Ro} $} & Avg \\
\midrule
\multirow{3}{*}{Small}   & FT      & 7.2	&24.1	&20.1&	17.1	
         \\
& AD     & 2.9&	25.1&	9.7&	12.6
         \\
& PHA    & 7.6&	24.9	&19.7	&\textbf{17.4}	
        \\
\midrule
\multirow{3}{*}{Base}      & FT      & 10.9	&25.3	&24.2	&20.1	
         \\
& AD     & 8.2	&26.5	&19.5	&18.1	
         \\
& PHA     & 10.8&	26.6	&24.8&	\textbf{20.7}	
         \\ 
\midrule
\multirow{3}{*}{Large} & FT      & 12.3&	24.7	&29.7	&22.2	
         \\
& AD     & 12.6	&27.1	&25.9	&21.9
         \\
& PHA     &14.2	&27.0	&28.6&	\textbf{23.3}	
   \\
\bottomrule
\end{tabular}
% \endgroup
}

\caption{\label{t10}
Result for NLU task. We report the ROUGE2 score for XSUM and BLEU score for Ro-En/En-Ro. The best result on each block is in \textbf{bold}. 
}
\end{table}

\end{document}